# Learning to Filter Spam E-Mail: A Comparison of a Naive Bayesian and a Memory-Based Approach[1]

Ion Androutsopoulos♣, Georgios Paliouras♣, Vangelis Karkaletsis♣, Georgios Sakkis♦, Constantine D. Spyropoulos♣ and Panagiotis Stamatopoulos♦

♣Software and Knowledge Engineering Laboratory
Institute of Informatics and Telecommunications
National Centre for Scientific Research "Demokritos"
153 10 Ag. Paraskevi, Athens, Greece
e-mail: {ionandr, paliourg, vangelis, costass}@iit.demokritos.gr

♦Department of Informatics, University of Athens
TYPA Buildings, Panepistimiopolis, 157 71 Athens, Greece
e-mail: {stud0926, T.Stamatopoulos}@di.uoa.gr

## Abstract

We investigate the performance of two machine learning algorithms in the context of anti-spam filtering. The increasing volume of unsolicited bulk e-mail (spam) has generated a need for reliable anti-spam filters. Filters of this type have so far been based mostly on keyword patterns that are constructed by hand and perform poorly. The Naive Bayesian classifier has recently been suggested as an effective method to construct automatically anti-spam filters with superior performance. We investigate thoroughly the performance of the Naive Bayesian filter on a publicly available corpus, contributing towards standard benchmarks. At the same time, we compare the performance of the Naive Bayesian filter to an alternative memory-based learning approach, after introducing suitable cost-sensitive evaluation measures. Both methods achieve very accurate spam filtering, outperforming clearly the keyword-based filter of a widely used e-mail reader.

## 1. Introduction

Electronic mail is an efficient and increasingly popular communication medium. Like every powerful medium, however, it is prone to misuse. One such case of misuse is the blind posting of unsolicited e-mail messages, also known as *spam*, to very large numbers of recipients. Spam messages are typically sent using bulk-mailers and address lists harvested from web pages and newsgroup archives. They vary significantly in content, from vacation advertisements to get-rich schemes. The common feature of these messages is that they are usually of little interest to the majority of the recipients. In some cases, they may even be harmful, e.g. spam messages advertising pornographic sites may be read by children. Apart from wasting time and bandwidth, spam e-mail also costs money to users with dial-up connections. A 1997 study (Cranor & Lamacchia 1998) reported that spam messages constituted approximately 10% of the incoming messages to a corporate network. The situation seems to be worsening, and without appropriate counter-measures, spam messages could eventually undermine the usability of e-mail.

---

Attempts to introduce legal measures against spam mailing have had limited effect.[2] A more effective solution is to develop tools to help recipients identify or remove automatically spam messages. Such tools, called *anti-spam filters*, vary in functionality from blacklists of frequent spammers to content-based filters. The latter are generally more powerful, as spammers often use fake addresses. Existing content-based filters search for particular keyword patterns in the messages. These patterns need to be crafted by hand, and to achieve better results they need to be tuned to each user and to be constantly maintained (Cranor & Lamacchia 1998), a tedious task, requiring expertise that a user may not have.

We address the issue of anti-spam filtering with the aid of machine learning. We examine supervised learning methods, which learn to identify spam e-mail after receiving training on messages that have been manually classified as spam or non-spam (hereafter *legitimate*). Learning algorithms of this type have been applied to several text categorization tasks (e.g. Apte & Damerau 1994, Lewis 1996, Dagan et al. 1997), including classifying e-mail into folders (Cohen 1996, Payne & Edwards 1997), or identifying interesting news articles (Lang 1995; see also Spertus 1997). Recently, Sahami et al. (1998) trained a Naive Bayesian classifier (Duda & Hart 1973, Mitchell 1997) for anti-spam filtering, reporting impressive performance on unseen messages. To our knowledge, this is the only previous attempt to apply machine learning to anti-spam filtering.

We have constructed a new benchmark corpus, which is a mixture of spam messages and messages sent via a moderated (and, hence, spam-free) mailing list. The corpus is made publicly available for other researchers to use as a benchmark.[3] Using this corpus, we performed a thorough evaluation of the Naive Bayesian algorithm, used in (Sahami et al. 1998), after introducing new cost-sensitive evaluation metrics. These are necessary to get an objective picture of the performance of the algorithm, when the cost of misclassification differs for the two classes (spam and legitimate). Furthermore, we used 10-fold cross-validation to get a more unbiased performance estimate, and investigated the effect of attribute-set size, an issue that had not been examined in (Sahami et al. 1998).

Another important contribution of the work presented here is the comparison of the Naive Bayesian classifier with another learning method, namely the memory-based classifier of TiMBL (Daelemans et al. 1999). We chose a memory-based classifier on the grounds that spam messages cover a very broad range of topics. This suggests that memory-based algorithms, that attempt to classify messages by finding similar previously received messages, may perform equally well as algorithms that attempt to learn unifying characteristics of spam messages. Our results confirmed this suspicion, and TiMBL achieved high classification accuracy. On average, the two learning methods performed equally well, with the best method depending on the exact usage scenario of the filter. Both methods outperformed clearly the keyword-based filter of Outlook 2000, a widely used e-mail reader.[4][5]

The remainder of this paper is organized as follows: section 2 describes our benchmark corpus; section 3 discusses preprocessing steps that are needed before applying the learning algorithms; section 4 presents the learning algorithms that we used; section 5 introduces cost-sensitive evaluation measures; section 6 discusses our experimental results; and section 7 concludes.

---

[2] Consult http://www.cauce.org, http://spam.abuse.net, and http://www.junkemail.org .

[3] The corpus is available from the publications section of http://www.iit.demokritos.gr/~ionandr .

[4] "Outlook 2000" is a trademark of Microsoft Corporation. Outlook's documentation points to a file containing the patterns of its anti-spam filter. We tried both a case-sensitive and a case-insensitive version of these patterns, and use the best-performing version in each experiment.

[5] An earlier summary of our experiments with the Naive Bayesian classifier, not comprising the experiments with TiMBL and Outlook 2000, can be found in (Androutsopoulos *et al.* 2000a).

## 2. Corpus collection

The benchmark corpus that we constructed is a mixture of spam messages and messages received via the Linguist list, a moderated mailing list about the profession and science of linguistics.[6] The corpus, dubbed *Ling-Spam*, consists of 2893 messages:

- 2412 Linguist messages, obtained by randomly downloading digests from the list's archives, breaking the digests into their messages, and removing text added by the list's server.

- 481 spam messages, received by the first author. Attachments, HTML tags, and duplicate spam messages received on the same day were not included.

Spam messages are 16.6% of the corpus, a figure close to the incoming spam rates of the authors, and rates reported in (Sahami *et al.* 1998) and (Cranor & LaMacchia 1998).

Although the Linguist messages are more topic-specific than most users' incoming e-mail, they are less standardized than one might expect (e.g. they contain job postings, software availability announcements, even flame-like responses). Hence, useful preliminary conclusions about anti-spam filtering can be reached with Ling-Spam, until better public corpora become available.[7] With a more direct interpretation, our experiments can also be seen as a study on anti-spam filters for open un-moderated mailing lists or newsgroups.

## 3. Corpus preprocessing

For every message in Ling-Spam, a vector representation $\vec{x} = \langle x_1, x_2, x_3, \ldots, x_n \rangle$ was computed, where $x_1, \ldots, x_n$ are the values of attributes $X_1, \ldots, X_n$, much as in the vector space model (Salton & McGill 1983). Following (Sahami *et al.* 1998), all attributes are binary: $X_i = 1$ if some characteristic represented by $X_i$ is present in the message; otherwise $X_i = 0$. In our experiments, attributes correspond to words, i.e. each attribute shows if a particular word (e.g. "adult") occurs in the message. It is also possible, however, to introduce attributes corresponding to phrases (e.g. showing if "be over 21" is present) or non-textual properties (e.g. whether or not a message contains attachments; see Sahami *et al.* 1998).

As in (Sahami *et al.* 1998), to select among all possible attributes (in our case, all possible word-attributes), we compute the mutual information ($MI$) of each candidate attribute $X$ with the category-denoting variable $C$:

$$MI(X;C) = \sum_{x \in \{0,1\}, c \in \{spam, legit\}} P(X=x, C=c) \cdot \log \frac{P(X=x, C=c)}{P(X=x) \cdot P(C=c)}$$

The attributes with the $m$ highest $MI$-scores are then selected. The probabilities are estimated from the training corpus as frequency ratios.[8] To avoid treating forms of the same word as different attributes, a lemmatizer was applied to Ling-Spam, substituting each word by its base form (e.g. "earning" becomes "earn").[9]

---

[6] The Linguist list is archived at http://listserv.linguistlist.org/archives/linguist.html .
[7] To address privacy issues, we have recently started experimenting with suitably "encoded" personal e-mail folders. Consult (Androutsopoulos *et al.* 2000b).
[8] Consult (Mitchell 1996) for more elaborate estimates.
[9] We used *morph*, a lemmatizer included in GATE. See http://www.dcs.shef.ac.uk/research/groups/nlp/gate .

## 4. Classification of e-mail messages

We now turn to the learning algorithms we experimented with.

### 4.1. Naive Bayesian classification

From Bayes' theorem and the theorem of total probability, the probability that a document $d$ with vector $\vec{x} = \langle x_1, \ldots, x_n \rangle$ belongs to category $c$ is:

$$P(C = c \mid \vec{X} = \vec{x}) = \frac{P(C = c) \cdot P(\vec{X} = \vec{x} \mid C = c)}{\sum_{k \in \{spam, legit\}} P(C = k) \cdot P(\vec{X} = \vec{x} \mid C = k)}$$

In practice, the probabilities $P(\vec{X} \mid C)$ are impossible to estimate without simplifying assumptions, because the possible values of $\vec{X}$ are too many and there are also data sparseness problems. The Naive Bayesian classifier assumes that $X_1, \ldots, X_n$ are conditionally independent given the category $C$, which yields:

$$P(C = c \mid \vec{X} = \vec{x}) = \frac{P(C = c) \cdot \prod_{i=1}^{n} P(X_i = x_i \mid C = c)}{\sum_{k \in \{spam, legit\}} P(C = k) \cdot \prod_{i=1}^{n} P(X_i = x_i \mid C = k)}$$

$P(X_i \mid C)$ and $P(C)$ are easy to estimate from the frequencies of the training corpus. A large number of empirical studies have found the Naive Bayesian classifier to be surprisingly effective (Langley *et al.* 1992, Domingos & Pazzani 1996), despite the fact that the assumption that the independence assumption is usually overly simplistic.[10]

Mistakenly blocking a legitimate message (classifying a legitimate message as spam) is generally more severe an error than letting a spam message pass the filter (classifying a spam message as legitimate). Let *legit* $\to$ *spam* and *spam* $\to$ *legit* denote the two error types. Invoking a decision-theoretic notion of cost, we assume that *legit* $\to$ *spam* is $\lambda$ times more costly than *spam* $\to$ *legit*. A message is classified as spam if the following criterion is met:

$$\frac{P(C = spam \mid \vec{X} = \vec{x})}{P(C = legitimate \mid \vec{X} = \vec{x})} > \lambda$$

To the extent that the independence assumption holds and the probability estimates are accurate, a classifier based on this criterion achieves optimal results (Duda & Hart 1973). In our case, $P(C = spam \mid \vec{X} = \vec{x}) = 1 - P(C = legitimate \mid \vec{X} = \vec{x})$, and the classification criterion is equivalent to:

$$P(C = spam \mid \vec{X} = \vec{x}) > t, \text{ with } t = \frac{\lambda}{1 + \lambda}, \lambda = \frac{t}{1 - t}$$

In the experiments of (Sahami *et al.* 1998), $t$ was set to 0.999, which corresponds to $\lambda = 999$. That is, mistakenly blocking a legitimate message was taken to be as bad as letting 999 spam messages pass the filter. When blocked messages are discarded without further processing, setting $\lambda$ to such a high value is reasonable, because in that case most users would consider losing a legitimate message unacceptable. Alternative usage scenarios are possible, however, and lower $\lambda$ values are reasonable in those cases.

---
[10] Consult (Friedman *et al.* 1997) for Bayesian classifiers with less restrictive independence assumptions.

For example, rather than being deleted, a blocked message could be returned to the sender, with an automatically inserted apology paragraph. The extra paragraph would explain that a filter blocked the message, and it would ask the sender to repost the message to a different, private un-filtered e-mail address of the recipient (see also Hall 1998). The private address would never be advertised (e.g. on web pages or newsgroups), making it unlikely to receive spam mail directly. Furthermore, the apology paragraph could include a frequently changing riddle (e.g. "Include in the subject the capital of France.") to ensure that spam messages are not forwarded automatically to the private address by robots that scan returned messages for new e-mail addresses. Messages sent to the private address without the correct riddle answer would be deleted automatically. (Spammers cannot afford the time to answer thousands of riddles.)

In the scenario of the previous paragraph, $\lambda = 9$ ($t = 0.9$) seems more reasonable: blocking a legitimate message is penalized mildly more than letting a spam message pass, to account for the fact that recovering from a blocked legitimate message requires overall more work (counting the sender's extra work to repost it) than recovering from a spam message that passed the filter (deleting it manually).

A third scenario would be to assume that the anti-spam filter simply flags messages it considers to be spam, without removing them from the user's mailbox (e.g. to help the user prioritize the reading of the messages). In that case, $\lambda = 1$ ($t = 0.5$) seems reasonable, since none of the two error types is significantly graver than the other.

### 4.2. Memory-based classification

The second method that we evaluated belongs to the family of memory-based (or instance-based) methods (Mitchell, 1997). The common feature of these methods is that they store *all* training instances in a memory structure, and use them directly for classification. The simplest form of memory structure is the multi-dimensional space defined by the attributes in the instance vectors. Each training instance is represented as a point in that space. The classification procedure is usually a variant of the simple *k*-nearest-neighbor (*k*-nn) algorithm. *k*-nn assigns to each new unseen instance the majority class among the *k* training instances that are closest to the unseen instance (its *k-neighborhood*).

We used the memory-based classification algorithm implemented in the TiMBL software (Daelemans *et al.*, 1999). TiMBL provides a basic memory-based classification algorithm and extensions to address issues such as efficient computation of the *k*-neighborhood and attribute weighting. We only used the basic algorithm, which is a variant of *k*-nn. One important difference from *k*-nn is in the definition of the *k*-neighborhood. TiMBL considers all the training instances at the *k* closest *distances* from the unseen instance. If there are more than one neighbors at each distance, the algorithm examines many more than *k* neighbors. In such cases, a small value of *k* is necessary, to avoid considering instances that are very different from the unseen one. A further addition we made to the basic TiMBL algorithm is a post-processing stage to take $\lambda$ into account. This simply multiplies the number of legitimate neighbors by $\lambda$, before deciding on the majority class in the neighborhood.

## 5. Measures to evaluate classification performance

In classification tasks, performance is often measured in terms of *accuracy* ($Acc$) or *error rate* ($Err = 1 - Acc$). Let $N_{legit}$ and $N_{spam}$ be the total numbers of legitimate and spam messages, respectively, to be classified by the filter, and $n_{Y \to Z}$ the number of messages belonging to category $Y$ that the filter classified as belonging to category $Z$ ($Y, Z \in \{legit, spam\}$). Then:

$$Acc = \frac{n_{legit \to legit} + n_{spam \to spam}}{N_{legit} + N_{spam}} \qquad Err = \frac{n_{legit \to spam} + n_{spam \to legit}}{N_{legit} + N_{spam}}$$

Accuracy and error rate assign equal weights to the two error types ($legit \to spam$ and $spam \to legit$). However, $legit \to spam$ is $\lambda$ times more costly than $spam \to legit$. To make accuracy and error rate sensitive to this cost difference, each legitimate message is treated, for evaluation purposes, as if it were $\lambda$ messages. That is, when a legitimate message is blocked, this counts as $\lambda$ errors; and when it passes the filter, this counts as $\lambda$ successes. This leads to the following definitions of *weighted accuracy* ($WAcc$) and *weighted error rate* ($WErr = 1 - WAcc$):

$$WAcc = \frac{\lambda \cdot n_{legit \to legit} + n_{spam \to spam}}{\lambda \cdot N_{legit} + N_{spam}} \qquad WErr = \frac{\lambda \cdot n_{legit \to spam} + n_{spam \to legit}}{\lambda \cdot N_{legit} + N_{spam}}$$

The values of accuracy and error rate (or their weighted versions) are often misleadingly high. To get a clear picture of a classifier's performance, it is common to compare its accuracy or error rate to those of a simplistic "baseline" approach. We use the case where no filter is present as our baseline: legitimate messages are (correctly) never blocked, and spam messages (mistakenly) always pass. The weighted accuracy and weighted error rate of the baseline are:

$$WAcc^b = \frac{\lambda \cdot N_{legit}}{\lambda \cdot N_{legit} + N_{spam}} \qquad WErr^b = \frac{N_{spam}}{\lambda \cdot N_{legit} + N_{spam}}$$

The *total cost ratio* ($TCR$) allows the performance of a filter to be compared easily to that of the baseline:

$$TCR = \frac{WErr^b}{WErr} = \frac{N_{spam}}{\lambda \cdot n_{legit \to spam} + n_{spam \to legit}}$$

Greater $TCR$ values indicate better performance. For $TCR < 1$, the baseline (not using the filter) is better. If cost is proportional to wasted time, an intuitive meaning for $TCR$ is the following: it measures how much time is wasted to delete manually all spam messages when no filter is used ($N_{spam}$), compared to the time wasted to delete manually any spam messages that passed the filter ($n_{spam \to legit}$) plus the time needed to recover from mistakenly blocked legitimate messages ($\lambda \cdot n_{legit \to spam}$).

For the benefit of readers more familiar with information retrieval and extraction tasks, our experimental results are also presented in terms of *spam recall* ($SR$) and *spam precision* ($SP$):

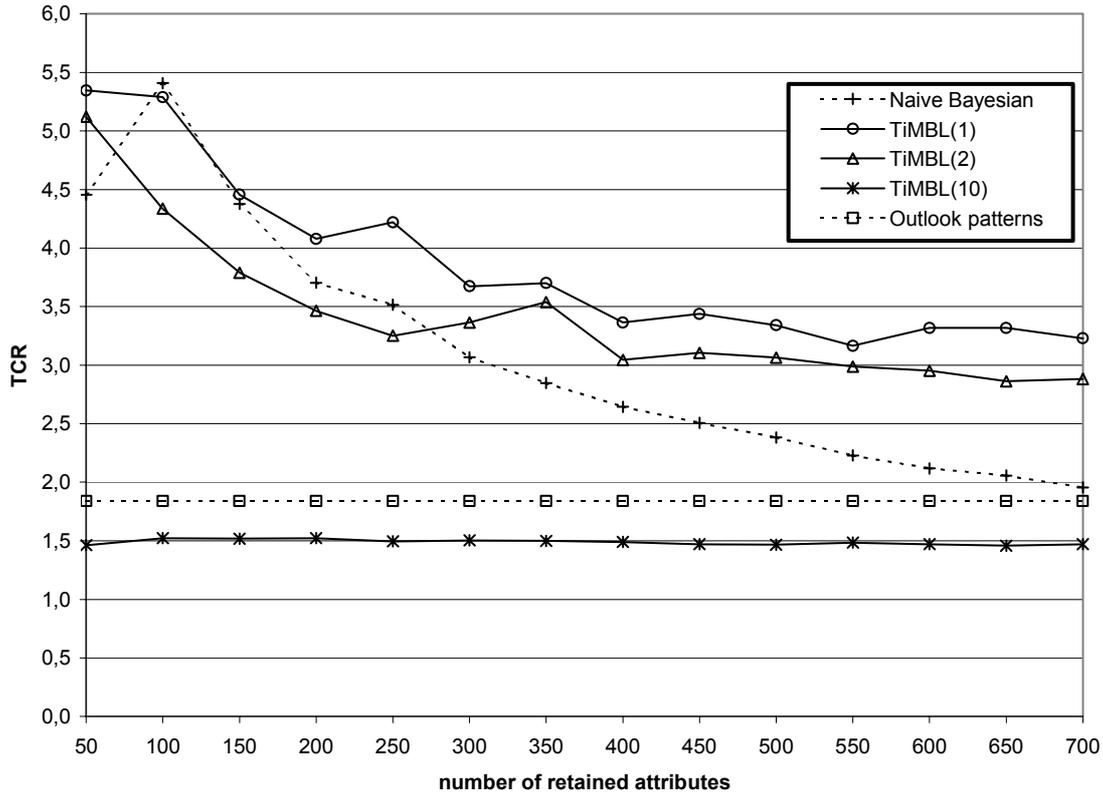

Figure 1. TCR scores for $\lambda=1$

$$SR = \frac{n_{spam \to spam}}{N_{spam}} \qquad SP = \frac{n_{spam \to spam}}{n_{spam \to spam} + n_{legit \to spam}}$$

Spam recall measures the percentage of spam messages that the filter manages to block (intuitively its effectiveness), while spam precision measures the degree to which the blocked messages are indeed spam (the filter's safety). Despite their intuitiveness, it is difficult to compare the performance of different filters using spam recall and precision: each filter (or filter configuration) yields a pair of spam recall and precision results; without a single unifying measure, like *TCR* that incorporates the cost difference between the two error types, it is difficult to decide which pair is better.[11]

## 6. Experimental results

We performed three sets of experiments on Ling-Spam, corresponding to the three scenarios (parameter $\lambda$) that were described in section 4.1. In each scenario, we varied the number of selected attributes from 50 to 700 by 50, each time retaining the attributes with the highest MI scores. *10-fold cross-validation* was used in all experiments: Ling-Spam was partitioned randomly into ten parts, and the experiment was repeated ten times, each time reserving a different part for testing, and using the remaining nine parts for training. *WAcc* was then averaged over the ten iterations, and *TCR* was computed as $WErr^b$ divided by the average *WErr*. The figures below show the average performance of each method in each experiment, including the *TCR* scores we obtained with Outlook's patterns. At the end of this section we select the best-performing configuration for each filter and scenario, and perform tests to establish statistically significant differences.

---

[11] The F-measure, used in information retrieval and extraction to combine recall and precision (e.g. Riloff & Lehnert 1994), cannot be used here, because its weighting factor cannot be related to the cost difference of the two error types.

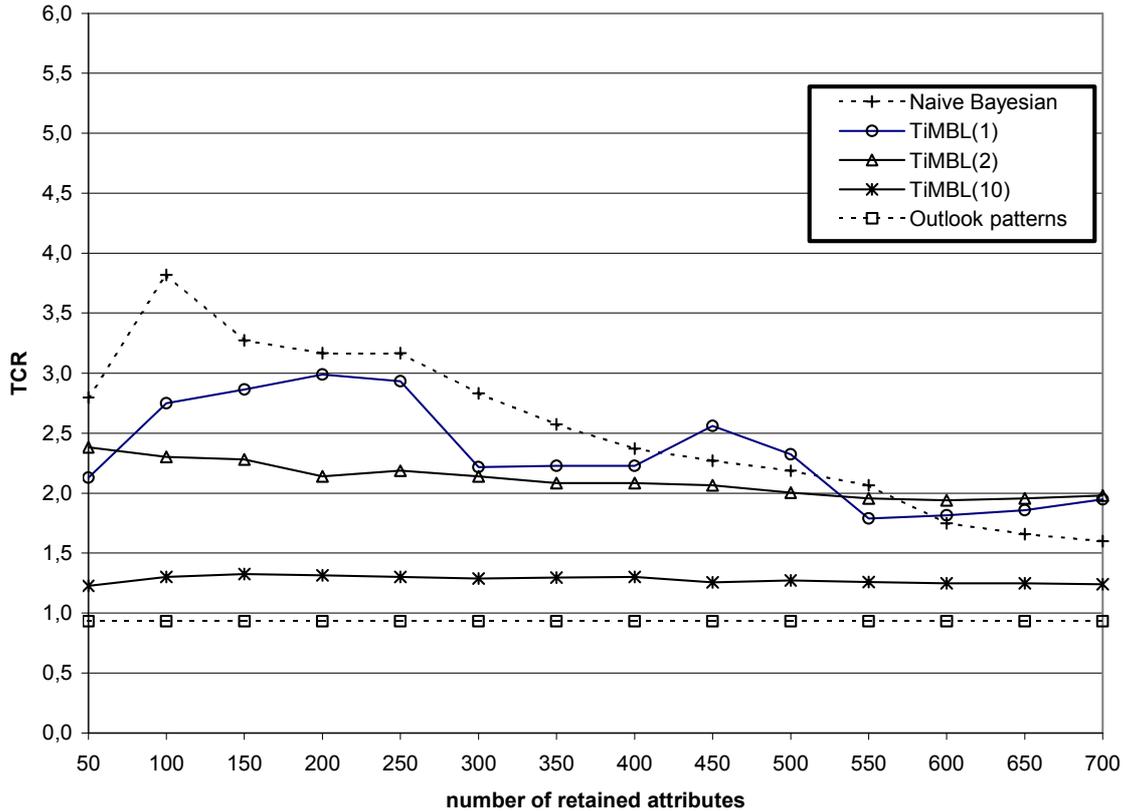

Figure 2. TCR scores for $\lambda$=9

### 6.1. Scenario 1: Flagging spam messages ($\lambda$=1)

In this scenario, the misclassification cost is identical for both error types. Figure 1 shows the corresponding results. The most important finding here is that both learning methods achieve very accurate classification, improving significantly on the baseline. Both methods perform better with small numbers of attributes. Their performance deteriorates as the size of the attribute set increases, which is due to the known sensitivity of the methods to data sparseness, caused by increasing the number of attributes.

The Naive Bayesian classifier performs best for 100 attributes, while TiMBL does best with the smallest attribute set size (50). TiMBL's performance was evaluated for three different values of $k$ (1,2,10). The method seems to perform best for small $k$ values. For $k=10$, the performance of the method falls to a very low level, improving only slightly on the base case. This is due to the large number of ties for each of the $k$ ($=10$) distances, which leads to a very large neighborhood (> 500 neighbors). In such cases, the behavior of the classifier approximates that of the default rule, which classifies everything according to the majority class (legitimate in our case). This is also responsible for the insensitivity of the method to the number of attributes for $k=10$. Outlook's keyword patterns perform very poorly compared to the other two methods, with the exception of TiMBL for $k=10$, which does even worse.

### 6.2. Scenario 2: Notifying senders about blocked messages ($\lambda$=9)

Here we increased the cost of misclassifying legitimate messages, by setting $\lambda = 9$. Figure 2 shows the corresponding results. Comparing to the previous scenario, the most important difference is the lower improvement of the learning methods over the baseline. This is due to the increased performance of the baseline as $\lambda$ increases: without a filter all legitimate messages are retained, and this becomes beneficial as $\lambda$ increases, making it harder to "beat" the baseline. The two methods also seem to be less sensitive to the size of the attribute set for $\lambda = 9$. This can be explained by the fact that after a certain number of attributes, the classification performance approaches its lowest possible value asymptotically. Another interesting observation is that the performance of Outlook's patterns falls below the base case, i.e. one is better off not using the filter.

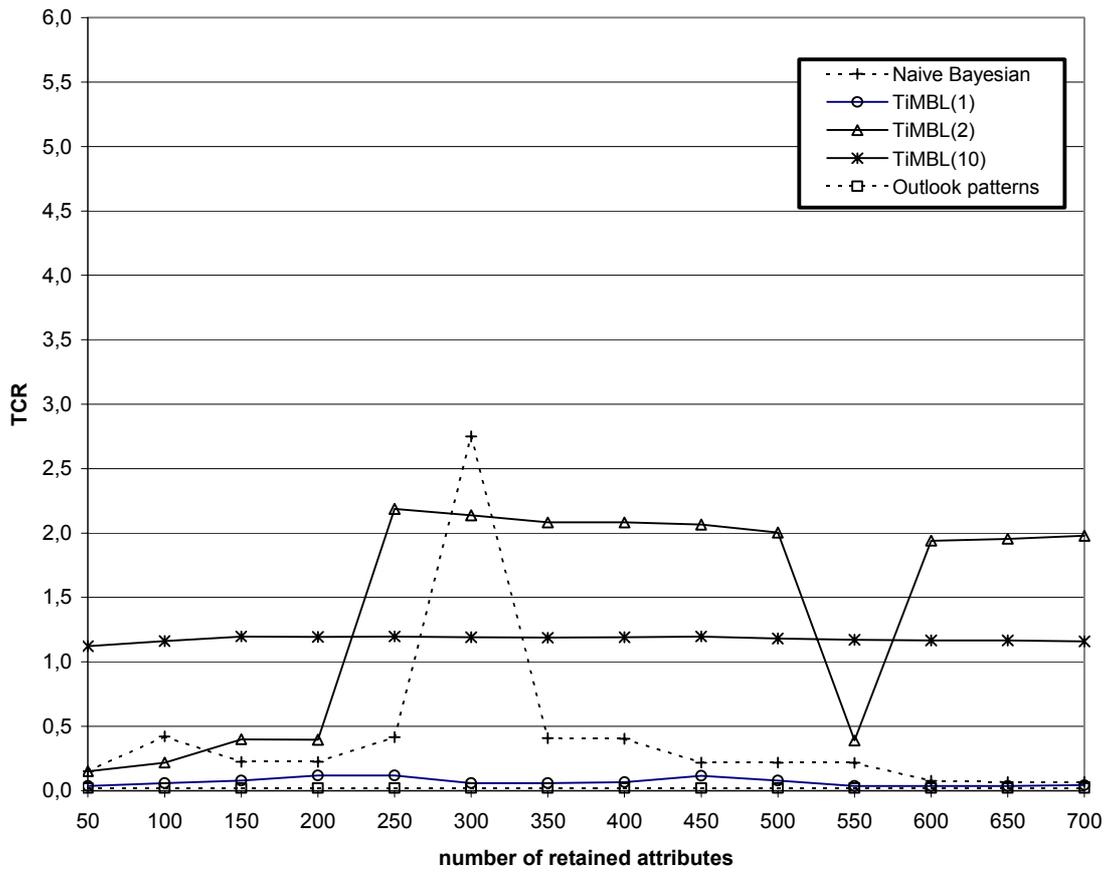

Figure 3. TCR scores for $\lambda$=999

## 6.3. Scenario 3: Removing blocked messages ($\lambda$=999)

In the third scenario a large $\lambda$ value is used (999). In this case, the choice to use any filter at all becomes doubtful, as the performance of the baseline increases to a level that any improvement on it is very hard. It is worth noting that this $\lambda$ value was the one used in (Sahami *et al.*, 1998). Figure 3 presents the scenario's results.

As expected, now all methods have difficulties achieving better results than the baseline. One exception is TiMBL for $k = 10$, which is consistently higher than the base case by a small margin. This is again an effect of the very large neighborhood, which now classifies most messages as legitimate, due to the large value of $\lambda$. The only instances that are classified as spam are those lying in an area of the instance space that is solely occupied by spam training instances, i.e. the most certain cases of unseen spam messages. TiMBL for $k = 2$, manages to achieve a significant improvement over the baseline for several consecutive attribute set sizes. Although the Naive Bayesian classifier achieves better performance for 300 attributes, this is the only point where it improves over the baseline. In practical applications, pinpointing the optimal attribute set size is infeasible, and hence TiMBL for $k=2$ is to be preferred. The reason for the abrupt fluctuations in the performance of the methods is that a single misclassification of a legitimate message causes a very large fall in TCR. This happens, for example, with TiMBL for $k=10$ and 550 attributes.

### 6.4. Best-performing configurations

Having investigated the effect of attribute set size, we now concentrate on the attribute set sizes for which each learning method performs best, and examine whether the differences between the methods are statistically significant. Table 1 presents the results for each method and $\lambda$ value by decreasing *TCR*. Paired single-tailed t-tests on *WAcc* show that the performance differences between the filter configurations of table 1 are all statistically significant at $p < 0.05$, with the following exceptions (italics in table 1): Naive Bayesian and TiMBL ($k=1,2$) for $\lambda=1$; Naive Bayesian and TiMBL ($k=1$) for $\lambda=9$.

We note that for a corpus of similar spam rate, (Sahami *et al.* 1998) reports 92.3% spam precision and 80.0% spam recall using the Naive Bayesian classifier at 500 attributes and $\lambda = 999$. No principled comparison to these results can be made, however, as they were obtained using a different corpus and additional manually selected phrasal and non-textual attributes

## 7. Conclusions

We performed a thorough evaluation of two learning methods on the task of anti-spam filtering, using a corpus that we made publicly available, and suitable cost-sensitive evaluation measures. Both methods achieved very high classification accuracy and clearly outperformed the anti-spam keyword patterns of a widely used e-mail reader. Our findings suggest that it is entirely feasible to construct learning-based anti-spam filters when spam messages are simply to be flagged, or when additional mechanisms are available to inform the senders of blocked messages. When no such mechanisms are present, a memory-based approach appears to be more viable, but great care is needed to configure the filter appropriately.

| Filter used | $\lambda$ | no. of attributes | spam recall (%) | spam precision (%) | weighted accuracy (%) | TCR |
|---|---|---|---|---|---|---|
| *Naive Bayesian* | 1 | 100 | 82.35 | 99.02 | 96.926 | 5.41 |
| *TiMBL(1)* | | 50 | 85.27 | 95.92 | 96.890 | 5.35 |
| *TiMBL(2)* | | 50 | 83.19 | 97.10 | 96.753 | 5.12 |
| Outlook patterns | | – | 53.01 | 87.93 | 90.978 | 1.84 |
| TiMBL(10) | | 100 | 34.54 | 99.64 | 89.079 | 1.52 |
| Baseline (no filter) | | – | 0 | ∞ | 83.374 | 1 |
| *Naive Bayesian* | 9 | 100 | 77.57 | 99.45 | 99.432 | 3.82 |
| *TiMBL(1)* | | 200 | 74.05 | 98.97 | 99.274 | 2.99 |
| TiMBL(2) | | 50 | 63.66 | 99.05 | 99.090 | 2.38 |
| TiMBL(10) | | 150 | 24.55 | 100.00 | 98.364 | 1.33 |
| Baseline (no filter) | | – | 0 | ∞ | 97.832 | 1 |
| Outlook patterns | | – | 39.29 | 88.32 | 97.670 | 0.93 |
| Naive Bayesian | 999 | 300 | 63.67 | 100.00 | 99.993 | 2.86 |
| TiMBL(2) | | 250 | 54.30 | 100.00 | 99.991 | 2.22 |
| TiMBL(10) | | 250 | 16.45 | 100.00 | 99.983 | 1.18 |
| Baseline (no filter) | | – | 0 | ∞ | 99.980 | 1 |
| TiMBL(1) | | 200 | 74.05 | 98.97 | 99.829 | 0.12 |
| Outlook patterns | | – | 39.29 | 88.32 | 98.952 | 0.02 |

Table 1: Results on the Ling-Spam corpus using the best configurations

We are currently examining alternative learning methods for the same task, including attribute-weighted versions of the memory-based algorithm. We also plan to explore alternative attribute selection techniques, including term extraction methods to move from word to phrasal attributes.

## Acknowledgements

Part of this work was performed using text categorization machinery developed within the context of ADIET (Adaptive Information Extraction Technology), a bilateral cooperation project funded by the governments of Greece and France.